\documentclass[11pt,a4paper]{article}
\usepackage[hyperref]{acl2017}
\usepackage{times}
\usepackage{latexsym}
\usepackage{amssymb}
\usepackage{amsmath}
\usepackage{subfig}
\usepackage{paralist}
\usepackage{balance}

\usepackage{graphicx}
\usepackage{color}

\usepackage{url}
\usepackage{multirow}

\aclfinalcopy 
\def\aclpaperid{3463} 

\title{Chunk-Based Bi-Scale Decoder for Neural Machine Translation}

\author{Hao Zhou\thanks{~Work was done when Hao Zhou was interning and Zhaopeng Tu was working at Huawei Noah's Ark Lab.} \\ Nanjing University \\ {\normalsize \tt zhouh@nlp.nju.edu.cn} \And 
	      Zhaopeng Tu$^{*}$ \\ Tencent AI Lab \\ {\normalsize \tt tuzhaopeng@gmail.com} \And
	      Shujian Huang \\ Nanjing University \\ {\normalsize \tt huangsh@nlp.nju.edu.cn}
	      \AND
	      Xiaohua Liu \\ Huawei Noah's Ark Lab \\ {\normalsize \tt liuxiaohua3@huawei.com} \And
	      Hang Li	\\ Huawei Noah's Ark Lab \\ {\normalsize \tt hangli.hl@huawei.com} \And
	      Jiajun Chen	\\ Nanjing University \\ {\normalsize \tt chenjj@nlp.nju.edu.cn}}

\begin{document}
\maketitle

\begin{abstract}

In typical neural machine translation~(NMT), the decoder generates a sentence word by word, packing all linguistic granularities in the same time-scale of RNN. 
In this paper, we propose a new type of decoder
for NMT, which splits the decode state into two parts and updates them in two different time-scales.
Specifically, we first predict a chunk time-scale state for phrasal modeling, on top of which multiple word time-scale states are generated.
In this way, the target sentence is translated hierarchically from chunks to words, with information in different granularities being leveraged.
Experiments show that our
proposed model significantly improves the
translation performance over the state-of-the-art
NMT model.



\end{abstract}

\section{Introduction}

Recent work of neural machine translation~(NMT) models propose to adopt the \textit{encoder-decoder} framework for machine translation~\citep{D13-1176,D14-1179,sutskever2014sequence}, which employs a recurrent neural network (RNN) \textit{encoder} to model the source context information and a RNN \textit{decoder} to generate translations, which is significantly different from previous statistical machine translation systems~\citep{koehn2003statistical,chiang2005hierarchical}.
This framework is then extended by an attention mechanism, which acquires source sentence context dynamically at different decoding steps~\citep{bahdanau2014neural,D15-1166}.

The decoder state stores translation information at different granularities, determining which segment should be expressed~(phrasal), and which word should be generated~(lexical), respectively.
However, due to the extensive existence
of multi-word phrases and expressions, the varying speed of the lexical component is much faster than the phrasal one.
As in the generation of ``\textit{the French Republic}", the lexical component in the decoder will change thrice, each of which for a separate word. But the phrasal component may only change once.
The inconsistent varying speed of the two components may cause translation errors.

Typical NMT model generates target sentences in the word level, packing the phrasal and lexical information in one hidden state, which is not necessarily the best for translation.
Much previous work propose to improve the NMT model by adopting fine-grained translation levels such as the character or sub-word levels, which can learn the intermediate information inside words~\citep{ling2015character,P16-2058,chung-cho-bengio:2016:P16-1,luong2016iclr_multi,lee2016fully,sennrich-haddow:2016:WMT,P16-1162,garcia2016factored}.  
However, high level structures such as phrases has not been explicitly explored in NMT, which is very useful for machine translation~\citep{koehn2007moses}.


We propose a chunk-based bi-scale decoder for NMT, which explicitly splits the lexical and phrasal components into different time-scales.\footnote{In this work, we focus on chunk-based well-formed phrases, which generally contain two to five words.}
The proposed model generates target words in a hierarchical way, which deploys a standard word time-scale RNN (lexical modeling) on top of an additional chunk time-scale RNN (phrasal modeling).
At each step of decoding, our model first predict a chunk state with a {\em chunk attention}, based on which multiple word states are generated without attention. 
The word state is updated at every step, while the chunk state is only updated when the chunk boundary is detected by a \textit{boundary gate} automatically. In this way, we incorporate {\em soft phrases} into NMT, which makes the model flexible at capturing both {\em global reordering} of phrases and {\em local translation} inside phrases.
Our model has following benefits:

\begin{compactenum}
\item The chunk-based NMT model explicitly splits the lexical and phrasal components of the decode state for different time-scales, which addresses the issue of inconsistent updating speeds of different components, making the model more flexible.

\item Our model recognizes phrase structures explicitly.
Phrase information are then used for word predictions, the representations of which are then used to help predict corresponding words.

\item Instead of incorporating source side linguistic information~\citep{P16-1078,sennrich-haddow:2016:WMT}, our model incorporates linguistic knowledges in the target side (for deciding chunks), which will guide the translation more in line with linguistic grammars.

\item Given the predicted phrase representation, our NMT model could extract attentive source context by \textit{chunk attention}, which is more specific and thus more useful compared to the word-level counterpart.

\end{compactenum}

Experiments show that our proposed model obtains considerable BLEU score improvements upon an attention-based NMT baseline on the Chinese to English and the German to English datasets simultaneously.

\section{Standard Neural Machine Translation Model}

Generally, neural machine translation system directly models the conditional probability of the translation $y$ word by word~\citep{bahdanau2014neural}.
Formally, given an input sequence $\mathbf{x}$ $=$ [$x_1$, $x_2$, $\dots$ , $x_{J}$], and the previously generated sequence
$\mathbf{y_{<t}}$ = [$y_1$, $y_2$, $\dots$ , $y_{t-1}$], the probability of next target word $y_t$ is
\begin{equation}
P(y_{t} | \mathbf{x}) = softmax(f(e_{y_{t-1}}, s_t, c_t))
\label{eqn-probability}
\end{equation} 
where $f(\cdot)$ is a non-linear function, $e_{y_{t-1}}$ is the embedding of $y_{t-1}$; $s_t$ is the \textit{decode state} at the time step $t$, which is computed by
\begin{align}
s_t = g(s_{t-1}, e_{y_{t-1}}, c_t)
\label{eqn-decoder-state}
\end{align}
Here $g(\cdot)$ is a transition function of decoder RNN. 
$c_t$ is the context vector computed by
\begin{equation}
c_t = \sum_{j=1}^{J} \textsc{Att}(s_{t-1}, h_j) \cdot h_j = \sum_{j=1}^{J}\alpha_{t,j} \cdot h_j \label{equ:attention}
\end{equation}
where $\textsc{Att}$ is an attention operation, which outputs alignment distribution $\alpha$:
\begin{equation}
\alpha_{t,j} = \frac{exp(e_{t, j})}{\sum_{k = 1}^{T_x}exp(e_{t, k})}
\end{equation}
\begin{equation}
e_{t, j} = v_a^T \tanh(W_a{s_{t-1}} + U_ah_j )
\end{equation}
and $h$ is the annotation of
${\bf x}$ from a bi-directional RNNs.
The training objective is to maximize the likelihood of the training data.
Beam search is adopted for decoding.

\section{Chunk-Based Bi-Scale Neural Machine Translation Model}
\label{sec:model}

\begin{figure}
 \centering
 \includegraphics[scale=.35]{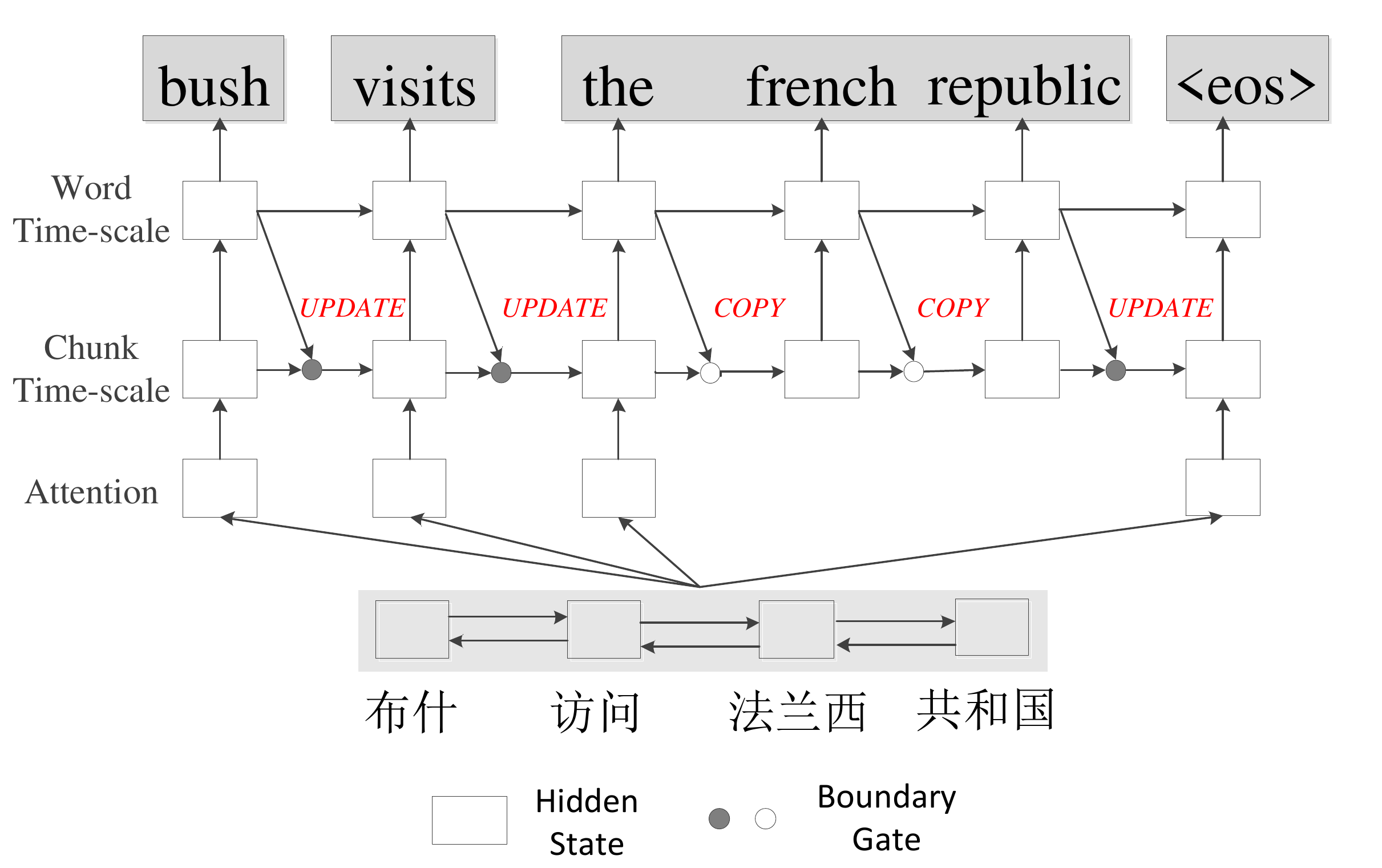}
   \caption{The architecture of the chunk-based bi-scale NMT.}
  \label{fig:multiscale}
\end{figure}

Instead of the word-based decoder, we propose to use a chunk-based bi-scale decoder, which generates translation hierarchically with {\it chunk} and {\it word} time-scales, as shown in Figure~\ref{fig:multiscale}.
Intuitively, we firstly generate a chunk state with the attention model, which extracts the source context for the current phrasal scope. Then we generate multiple lexical words based on the same chunk state, which does not require attention operations. The boundary of a chunk is determined by a {\em boundary gate}, which decides whether to update the chunk state or not at each step.

Formally, 
the probability of next word $y_t$ is
\begin{eqnarray}
P(y_{t} | \mathbf{x}) &=& softmax(f(e_{y_{t-1}}, s_t, p_t)) \\
s_t &=& g(s_{t-1}, e_{y_{t-1}}, p_t)
\end{eqnarray}
here $p_t$ is the chunk state at step $t$. Compared with Equations~\ref{eqn-probability} and~\ref{eqn-decoder-state}, the generation of target word is based on the chunk state instead of the context vector $c_t$ produced by the attention model.

Since a chunk may correspond to multiple words, we employ a {\em boundary gate} $b_t$ to decide the boundary of each chunk:
\begin{equation}
p(b_t) = softmax(s_{t-1}, e_{y_{t-1}})
\end{equation}
$b_t$ will be $0$ or $1$, where $1$ denotes this is the boundary of a new chunk while $0$ denotes not. Two different operations would be executed:
\begin{align}
p_t=&\begin {cases}
p_{t-1}, \quad\quad\quad\quad\quad\quad  b_t = 0~\textsc{(Copy)} \\
g(p_{t-1}, e_{p_{t-1}}, pc_t), ~~ b_t = 1~\textsc{(Update)}
\end{cases}\nonumber
\end{align}
In the \textsc{Copy} operation, the chunk state is kept the same as the previous step. 
In the \textsc{Update} operation, $e_{p_{t-1}}$ is the representation of last chunk, which is computed by the LSTM-minus approach \citep{wang2016graph}:
\begin{equation}
e_{p_{t-1}} = m(s_{t-1}, e_{y_{t-1}}) - {  m({s_{t'}}, e_{y_{t'}} } )
\end{equation} 
here $t'$ is the boundary of last chunk and $m(\cdot)$ is a linear function.
$pc_t$ is the context vector for chunk $p_t$, which is calculated by a {\em chunk attention} model:
\begin{equation}
pc_t = \sum_{j=1}^{T_s} \textsc{Att}(p_{t-1}, h_j) \cdot h_j
\end{equation}
The chunk attention model differs from the standard word attention model (i.e., Equation~\ref{equ:attention}) at: 1) it reads chunk state $p_{t-1}$ rather than word state $s_{t-1}$, and 2) it is only executed at boundary of each chunk rather than at each decoding step.

In this way, our model only extracts source context once for a chunk, and the words in one chunk will share the same context for word generation.
The chunk attention mechanism adds a constrain that target words in the same chunk shares the same source context.

\paragraph{Training}
To encourage the proposed model to learn reasonable chunk state, we add two additional objectives in training:

\noindent{\bf \em Chunk Tag Prediction}: For each chunk, we predict the probability of its tag $P(l_k | {\bf x}) = softmax\big(f(p_t, e_{p_{t}}, c_t)\big)$, where $l_k$ is the syntactic tag of the $k$-th chunk such as \textit{NP}~(noun phrase) and \textit{VP}~(verb phrase), and $t$ is time step of its boundary.

\noindent{\bf \em Chunk Boundary Prediction}: At each decoding step, we predict the probability of chunk boundary $P(b_t | {\bf x}) = softmax(s_{t-1}, e_{y_{t-1}})$.

Accordingly, given a set of training examples $\{\left[{\bf x}_n, {\bf y}_n\right]\}_{n=1}^{N}$, the new training objective is
\begin{eqnarray}
J(\theta, \gamma) = \arg\max \sum_{n=1}^{N} \bigg\{ \log P({\bf y_n}|{\bf x_n}) \nonumber\\
			+ \log P({\bf l_n}|{\bf x_n}) + \log P({\bf b_n}|{\bf x_n})  \bigg\}
\end{eqnarray}
where ${\bf l_n}$ and ${\bf b_n}$ are chunk tag sequence and boundary sequence on ${\bf y_n}$, respectively.

\section{Experiments}


We carry out experiments on a Chinese-English
translation task.
Our training data consists of 1.16M\footnote{3LDC2002E18, LDC2003E14, the Hansards portion of LDC2004T08, and LDC2005T06.} sentence pairs extracted from LDC corpora, with 25.1M Chinese words and 27.1M English words, respectively. 
We choose the NIST 2002 (MT02) dataset as our development set, and the NIST 2003 (MT03), 2004 (MT04) 2005 (MT05) datasets as our test sets.
We also evaluate our model on the
WMT translation task of German-English, newstest2014~(DE14) is adopted as development set and newstest2012, newstest2013~(DE1213) are adopted as testing set.
The English sentences are labeled by a neural chunker, which is implemented according to \citet{zhou-zhang-huang-chen:2015:ACL}.
We use the case-insensitive 4-gram NIST BLEU score as our evaluation metric~\citep{papineni2002bleu}.

In training, we limit the source and target vocabularies to the most
frequent 30K words.
We train each model with the sentences
of length up to 50 words.
Sizes of the chunk representation and chunk hidden state are set to 1000.
All the other settings are
the same as in \citet{bahdanau2014neural}.

\subsection{Results on Chinese-English}

\begin{table*}[tp]
\centering
\begin{tabular}{l|c|ccc|c}
\hline
{System} & MT02 & MT03 & MT04 & MT05 & Ave. \\
\hline
{Moses} &  30.10 & 28.82 & 31.22 & 27.78 &  29.48 \\
\hline
{dl4mt}  & 31.66 & 29.92 & 32.76 & 28.88 & 30.81 \\
{dl4mt-2}   & 31.01 & 28.74 & 31.71 & 27.95 & 29.85 \\
\hline
{This Work} & \textbf{33.43} & \textbf{32.06} & \textbf{34.21} & \textbf{30.01} & \textbf{32.42}\\
\hline

\end{tabular}
\caption{BLEU scores for different systems.}
\label{tab-final}
\end{table*}

\begin{table*}[]
\centering

\begin{tabular}{c|c|ccc|c}
\hline
 Attention & MT02 & MT03 & MT04 & MT05	& Ave.  \\
  \hline
 Word & 32.69 & 31.36 & 33.55& 29.77 & 31.56\\
 \hline
 Chunk & 33.43 & 32.06 & 34.21 & 30.01 & 32.42\\
 \hline
\end{tabular}
\caption{Results with different attention models.}
\label{tab_ablation}
\end{table*} 

We list the BLEU score of our proposed model in Table \ref{tab-final}, comparing with Moses~\citep{koehn2007moses} and dl4mt\footnote{\url{https://github.com/nyu-dl/dl4mt -tutorial}}~\citep{bahdanau2014neural}, which are state-of-the-art models of SMT and NMT, respective.
For Moses, we use the default
configuration with a 4-gram language model
trained on the target portion of the training data.
For dl4mt, we also report the results~(dl4mt-2) by using two decoder layers~\citep{wu2016google} for better comparison.

As shown in Table \ref{tab-final}, our proposed model outperforms different baselines on all sets, which verifies that the chunk-based bi-scale decoder is effective for NMT.
Our model gives a 1.6 BLEU score improvement upon the standard NMT baseline~(dl4mt).
We conduct experiment with dl4mt-2 to see whether the neural NMT system can model the bi-scale components with different varying speeds automatically.
Surprisingly, we find that dl4mt-2 obtains lower BLEU scores than dl4mt.
We speculate that the more complex model dl4mt-2 may need more training data for obtaining reasonable results.

\paragraph{Effectiveness of Chunk Attention}
As described in Section \ref{sec:model}, we propose to use the \textit{chunk attention} to replace the word level attention in our model, in which the source context extracted by the chunk attention will be used for the corresponding word generations in the chunk.
We also report the result of our model using conventional word attention for comparison.
As shown in Table \ref{tab_ablation}, our model with the chunk attention gives higher BLEU score than the word attention. 

\begin{figure}[]
 \centering
  \includegraphics[scale=.17]{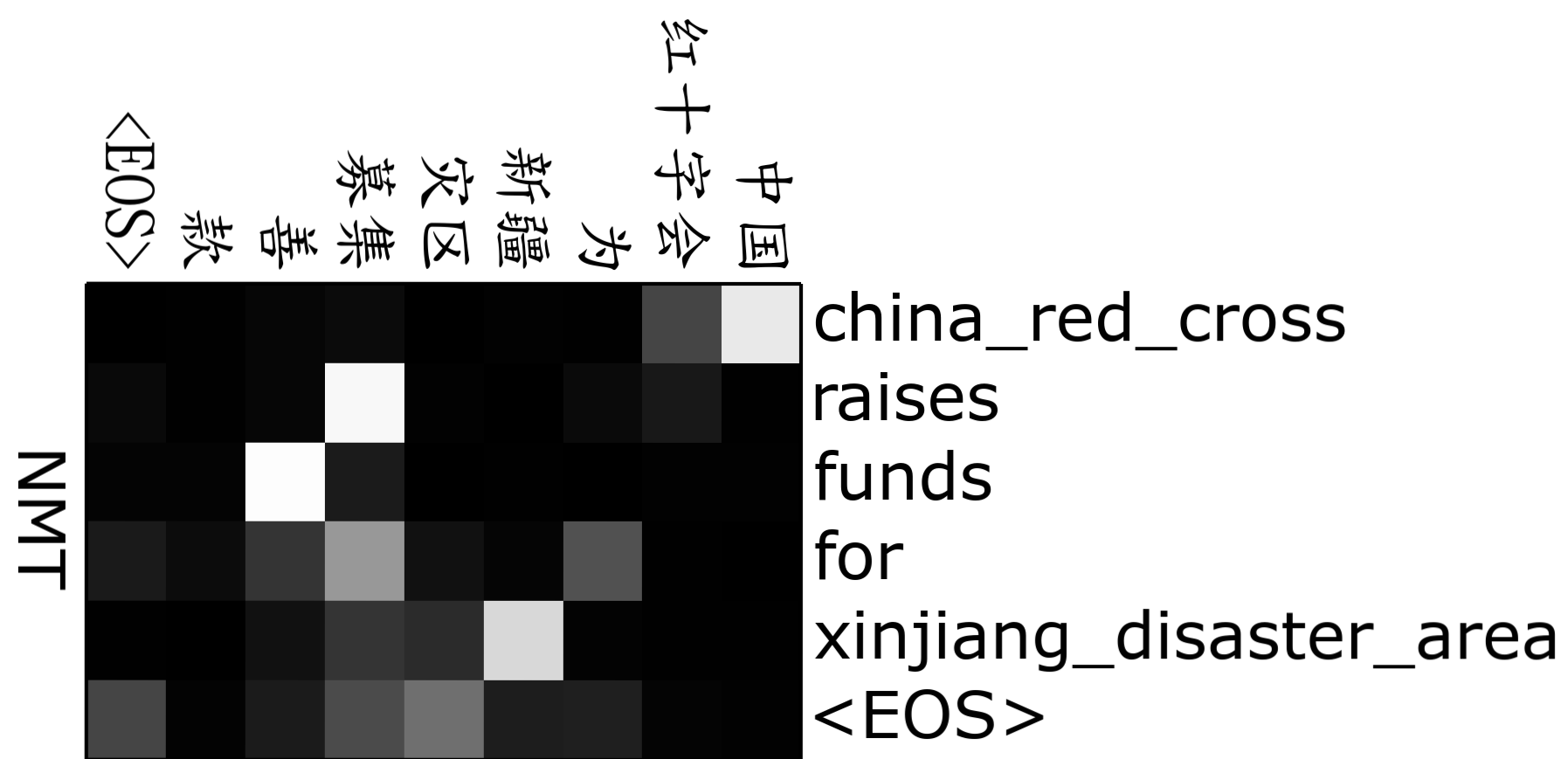}
   \caption{Alignments with chunk attention.}
  \label{fig:align}
\end{figure}

Intuitively, we think chunks are more specific in semantics, thus could extract more specific source context for translation.
The chunk attention could be considered as a compromise approach between encoding the whole source sentence into decoder without attention~\citep{sutskever2014sequence} and utilizing word level attention at each step~\citep{bahdanau2014neural}.
We also draw the figure of alignments by chunk attention~(Figure \ref{fig:align}), from which we can see that our chunk attention model can well explore the alignments from phrases to words.

\paragraph{Predictions of the Chunk Boundary and Chunk Label}
\begin{table}[h]
\centering
\begin{tabular}{c|cccc}
\hline
 & MT02 & MT03 & MT04 & MT05  \\
 \hline
 Boundary &89.97 & 88.81 & 89.64 & 89.25 \\
 \hline
 Label & 47.00 & 44.75 & 45.54 & 44.41  \\
 
 \hline
\end{tabular}
\caption{Accuracies of predicted chunk boundary and chunk label.}
\label{tab:boundary}
\end{table}

We also compute predicted accuracies of chunk boundaries and chunk labels on the auto-chunked development and testing data~(Table \ref{tab:boundary}).
We find that the chunk boundary could be predicted well, with an average accuracy of 89\%, which shows that our model could capture the phrasal boundary information in the translation process.
However, our model could not predict chunk labels as well as chunk boundaries. 
We speculate that more syntactic context features should be added to improve the performance of predicting chunk labels.

\paragraph{Subjective Evaluation}

\begin{table}[h]
\centering
\begin{tabular}{r|c|c}
\hline
 Model & dl4mt & Our Work \\
 \hline
 Adequacy &  3.26 & 3.35  \\
 Fluency &  3.69 &  3.71 \\
 Under-Translation& 50\% & 47\% \\
 Over-Translation &  32\% & 26\% \\
 \hline
\end{tabular}
\caption{Subjective evaluation results.}
\label{tab:sub}
\end{table}
Following \citet{P16-1008,Tu:2017:TACL,tu_etal:17}, we also compare our model with the dl4mt baseline by subjective evaluation.
Two human evaluators are asked to evaluate the translations of 100 source sentences randomly sampled from the test sets without knowing which system the  translation is translated by. 
The human evaluator is asked to give 4 scores:
adequacy score and fluency score, which are between 0 and 5, the larger, the better; under-translation score and over-translation score, which are set to 1 when under or over translation errors occurs, otherwise set to 0.

We list the averaged scores in Table \ref{tab:sub}.
We find that our proposed model improves the dl4mt baseline on both the translation adequacy and fluency aspects.
Specifically, the over translation error rate drops by 6\%, which confirms the assumption in the introduction that splitting the fast and slow varying components in different time-scales could help alleviate the over translation errors.

\subsection{Results on German-English}

\begin{table}[t]
\centering
\begin{tabular}{r|c|c}
\hline
 System & DE-14 & DE-1213 \\
 \hline
 dl4mt &  16.53 & 16.78  \\
 This Work &  17.40 &  17.45 \\
 \hline
\end{tabular}
\caption{Results on German-English}
\label{tab:sub}
\end{table}

We evaluate our model on the WMT’15 translation task from German to English.
We find that our proposed chunk-based NMT model also obtains considerable accuracy improvements on German-English.
However, the BLEU score gains are not as significant as on Chinese-English.
We speculate that the difference between Chinese and English is larger than German and English.
The chunk-based NMT model may be more useful for bilingual data with bigger difference.

\section{Related Work}

\paragraph{NMT with Various Granularities.}
A line of previous work propose to utilize other granularities besides words for NMT.
By further exploiting the character level~\citep{ling2015character,P16-2058,chung-cho-bengio:2016:P16-1,luong2016iclr_multi,lee2016fully}, or the sub-word level~\citep{sennrich-haddow:2016:WMT,P16-1162,garcia2016factored} information, the corresponding NMT models capture the information inside the word and alleviate the problem of unknown words.
While most of them focus on decomposing words into characters or sub-words, our work aims at composing words into phrases.

\paragraph{Incorporating Syntactic Information in NMT}
Syntactic information has been widely used in SMT~\citep{liu2006tree,marton2008soft,shen2008new}, and a lot of previous work explore to incorporate the syntactic information in NMT, which shows the effectiveness of the syntactic information~\citep{P16-2049}.
\citet{D16-1159} give some empirical results that the deep networks of NMT are able to capture some useful syntactic information implicitly.
~\citet{luong2016iclr_multi} propose to use a multi-task framework for NMT and neural parsing, achieving promising results. 
\citet{P16-1078} propose a string-to-tree NMT system by end-to-end training.
Different to previous work, we try to incorporate the syntactic information in the target side of NMT.
\citet{ishiwatari2017} concurrently propose to use chunk-based decoder to  cope with the problem of free word-order languages.
Differently, they adopt word-level attention, and predict the end of chunk by generating end-of-chunk tokens instead of using boundary gate.

\section{Conclusion}
We propose a chunk-based bi-scale decoder for neural machine translation, in which way, the target sentence is translated
hierarchically from chunks to words,
with information in different granularities
being leveraged.
Experiments show that our proposed model outperforms the standard attention-based neural machine translation baseline.
Future work includes abandoning labeled chunk data, adopting reinforcement learning to explore the boundaries of phrase automatically~\citep{mou2016coupling}.
Our code is released on \url{https://github.com/zhouh/chunk-nmt}. 

\section*{Acknowledge}
We would like to thank the anonymous reviewers
for their insightful comments. We also thank Lili Mou for helpful discussion and Hongjie Ji, Zhenting Yu, Xiaoxue Hou and Wei Zou for their help in data preparation and subjective evaluation.
This work was partially founded by the Natural Science
Foundation of China (61672277, 71503124) and
the China National 973 project 2014CB340301.

\balance
\bibliography{acl2017}

\begin{thebibliography}{}
\expandafter\ifx\csname natexlab\endcsname\relax\def\natexlab#1{#1}\fi

\bibitem[{Bahdanau et~al.(2014)Bahdanau, Cho, and Bengio}]{bahdanau2014neural}
Dzmitry Bahdanau, Kyunghyun Cho, and Yoshua Bengio. 2014.
\newblock Neural machine translation by jointly learning to align and
  translate.
\newblock In {\em ICLR 2015\/}.

\bibitem[{Chiang(2005)}]{chiang2005hierarchical}
David Chiang. 2005.
\newblock A hierarchical phrase-based model for statistical machine
  translation.
\newblock In {\em Proceedings of the 43rd Annual Meeting on Association for
  Computational Linguistics\/}. Association for Computational Linguistics,
  pages 263--270.

\bibitem[{Cho et~al.(2014)Cho, van Merrienboer, Gulcehre, Bahdanau, Bougares,
  Schwenk, and Bengio}]{D14-1179}
Kyunghyun Cho, Bart van Merrienboer, Caglar Gulcehre, Dzmitry Bahdanau, Fethi
  Bougares, Holger Schwenk, and Yoshua Bengio. 2014.
\newblock \href{https://doi.org/10.3115/v1/D14-1179}{Learning phrase
  representations using rnn encoder--decoder for statistical machine
  translation}.
\newblock In {\em Proceedings of the 2014 Conference on Empirical Methods in
  Natural Language Processing (EMNLP)\/}. Association for Computational
  Linguistics, pages 1724--1734.
\newblock
  \href{https://doi.org/10.3115/v1/D14-1179}{https://doi.org/10.3115/v1/D14-1179}.

\bibitem[{Chung et~al.(2016)Chung, Cho, and
  Bengio}]{chung-cho-bengio:2016:P16-1}
Junyoung Chung, Kyunghyun Cho, and Yoshua Bengio. 2016.
\newblock \href{http://www.aclweb.org/anthology/P16-1160}{A character-level
  decoder without explicit segmentation for neural machine translation}.
\newblock In {\em Proceedings of the 54th Annual Meeting of the Association for
  Computational Linguistics (Volume 1: Long Papers)\/}. Association for
  Computational Linguistics, Berlin, Germany, pages 1693--1703.
\newblock
  \href{http://www.aclweb.org/anthology/P16-1160}{http://www.aclweb.org/anthology/P16-1160}.

\bibitem[{Costa-juss{\`a} and Fonollosa(2016)}]{P16-2058}
R.~Marta Costa-juss{\`a} and R.~Jos{\'e}~A. Fonollosa. 2016.
\newblock \href{https://doi.org/10.18653/v1/P16-2058}{Character-based neural
  machine translation}.
\newblock In {\em Proceedings of the 54th Annual Meeting of the Association for
  Computational Linguistics (Volume 2: Short Papers)\/}. Association for
  Computational Linguistics, pages 357--361.
\newblock
  \href{https://doi.org/10.18653/v1/P16-2058}{https://doi.org/10.18653/v1/P16-2058}.

\bibitem[{Eriguchi et~al.(2016)Eriguchi, Hashimoto, and Tsuruoka}]{P16-1078}
Akiko Eriguchi, Kazuma Hashimoto, and Yoshimasa Tsuruoka. 2016.
\newblock \href{https://doi.org/10.18653/v1/P16-1078}{Tree-to-sequence
  attentional neural machine translation}.
\newblock In {\em Proceedings of the 54th Annual Meeting of the Association for
  Computational Linguistics (Volume 1: Long Papers)\/}. Association for
  Computational Linguistics, pages 823--833.
\newblock
  \href{https://doi.org/10.18653/v1/P16-1078}{https://doi.org/10.18653/v1/P16-1078}.

\bibitem[{Garc{\'\i}a-Mart{\'\i}nez et~al.(2016)Garc{\'\i}a-Mart{\'\i}nez,
  Barrault, and Bougares}]{garcia2016factored}
Mercedes Garc{\'\i}a-Mart{\'\i}nez, Lo{\"\i}c Barrault, and Fethi Bougares.
  2016.
\newblock Factored neural machine translation.
\newblock {\em arXiv preprint arXiv:1609.04621\/} .

\bibitem[{Ishiwatari et~al.(2017)Ishiwatari, Yao, Liu, Li, Zhou, Yoshinaga,
  Kitsuregawa, and Jia}]{ishiwatari2017}
Shonosuke Ishiwatari, Jingtao Yao, Shujie Liu, Mu~Li, Ming Zhou, Naoki
  Yoshinaga, Masaru Kitsuregawa, and Weijia Jia. 2017.
\newblock Chunk-based decoder for neural machine translation.
\newblock In {\em Proceedings of the 55th annual meeting of the Association for
  Computational Linguistics\/}. Association for Computational Linguistics.

\bibitem[{Kalchbrenner and Blunsom(2013)}]{D13-1176}
Nal Kalchbrenner and Phil Blunsom. 2013.
\newblock \href{http://aclweb.org/anthology/D13-1176}{Recurrent continuous
  translation models}.
\newblock In {\em Proceedings of the 2013 Conference on Empirical Methods in
  Natural Language Processing\/}. Association for Computational Linguistics,
  pages 1700--1709.
\newblock
  \href{http://aclweb.org/anthology/D13-1176}{http://aclweb.org/anthology/D13-1176}.

\bibitem[{Koehn et~al.(2007)Koehn, Hoang, Birch, Callison-Burch, Federico,
  Bertoldi, Cowan, Shen, Moran, Zens et~al.}]{koehn2007moses}
Philipp Koehn, Hieu Hoang, Alexandra Birch, Chris Callison-Burch, Marcello
  Federico, Nicola Bertoldi, Brooke Cowan, Wade Shen, Christine Moran, Richard
  Zens, et~al. 2007.
\newblock Moses: Open source toolkit for statistical machine translation.
\newblock In {\em Proceedings of the 45th annual meeting of the ACL on
  interactive poster and demonstration sessions\/}. Association for
  Computational Linguistics, pages 177--180.

\bibitem[{Koehn et~al.(2003)Koehn, Och, and Marcu}]{koehn2003statistical}
Philipp Koehn, Franz~Josef Och, and Daniel Marcu. 2003.
\newblock Statistical phrase-based translation.
\newblock In {\em Proceedings of the 2003 Conference of the North American
  Chapter of the Association for Computational Linguistics on Human Language
  Technology-Volume 1\/}. Association for Computational Linguistics, pages
  48--54.

\bibitem[{Lee et~al.(2016)Lee, Cho, and Hofmann}]{lee2016fully}
Jason Lee, Kyunghyun Cho, and Thomas Hofmann. 2016.
\newblock Fully character-level neural machine translation without explicit
  segmentation.
\newblock {\em arXiv preprint arXiv:1610.03017\/} .

\bibitem[{Ling et~al.(2015)Ling, Trancoso, Dyer, and Black}]{ling2015character}
Wang Ling, Isabel Trancoso, Chris Dyer, and Alan~W Black. 2015.
\newblock Character-based neural machine translation.
\newblock {\em arXiv preprint arXiv:1511.04586\/} .

\bibitem[{Liu et~al.(2006)Liu, Liu, and Lin}]{liu2006tree}
Yang Liu, Qun Liu, and Shouxun Lin. 2006.
\newblock Tree-to-string alignment template for statistical machine
  translation.
\newblock In {\em Proceedings of the 21st International Conference on
  Computational Linguistics and the 44th annual meeting of the Association for
  Computational Linguistics\/}. Association for Computational Linguistics,
  pages 609--616.

\bibitem[{Luong et~al.(2016)Luong, Le, Sutskever, Vinyals, and
  Kaiser}]{luong2016iclr_multi}
Minh-Thang Luong, Quoc~V. Le, Ilya Sutskever, Oriol Vinyals, and Lukasz Kaiser.
  2016.
\newblock Multi-task sequence to sequence learning.
\newblock In {\em International Conference on Learning Representations
  (ICLR)\/}. San Juan, Puerto Rico.

\bibitem[{Luong et~al.(2015)Luong, Pham, and Manning}]{D15-1166}
Thang Luong, Hieu Pham, and D.~Christopher Manning. 2015.
\newblock \href{https://doi.org/10.18653/v1/D15-1166}{Effective approaches to
  attention-based neural machine translation}.
\newblock In {\em Proceedings of the 2015 Conference on Empirical Methods in
  Natural Language Processing\/}. Association for Computational Linguistics,
  pages 1412--1421.
\newblock
  \href{https://doi.org/10.18653/v1/D15-1166}{https://doi.org/10.18653/v1/D15-1166}.

\bibitem[{Marton and Resnik(2008)}]{marton2008soft}
Yuval Marton and Philip Resnik. 2008.
\newblock Soft syntactic constraints for hierarchical phrased-based
  translation.
\newblock In {\em ACL\/}. pages 1003--1011.

\bibitem[{Mou et~al.(2016)Mou, Lu, Li, and Jin}]{mou2016coupling}
Lili Mou, Zhengdong Lu, Hang Li, and Zhi Jin. 2016.
\newblock Coupling distributed and symbolic execution for natural language
  queries.
\newblock {\em arXiv preprint arXiv:1612.02741\/} .

\bibitem[{Papineni et~al.(2002)Papineni, Roukos, Ward, and
  Zhu}]{papineni2002bleu}
Kishore Papineni, Salim Roukos, Todd Ward, and Wei-Jing Zhu. 2002.
\newblock Bleu: a method for automatic evaluation of machine translation.
\newblock In {\em Proceedings of the 40th annual meeting on association for
  computational linguistics\/}.

\bibitem[{Sennrich and Haddow(2016)}]{sennrich-haddow:2016:WMT}
Rico Sennrich and Barry Haddow. 2016.
\newblock \href{http://www.aclweb.org/anthology/W16-2209}{Linguistic input
  features improve neural machine translation}.
\newblock In {\em Proceedings of the First Conference on Machine
  Translation\/}. Association for Computational Linguistics, Berlin, Germany,
  pages 83--91.
\newblock
  \href{http://www.aclweb.org/anthology/W16-2209}{http://www.aclweb.org/anthology/W16-2209}.

\bibitem[{Sennrich et~al.(2016)Sennrich, Haddow, and Birch}]{P16-1162}
Rico Sennrich, Barry Haddow, and Alexandra Birch. 2016.
\newblock \href{https://doi.org/10.18653/v1/P16-1162}{Neural machine
  translation of rare words with subword units}.
\newblock In {\em Proceedings of the 54th Annual Meeting of the Association for
  Computational Linguistics (Volume 1: Long Papers)\/}. Association for
  Computational Linguistics, pages 1715--1725.
\newblock
  \href{https://doi.org/10.18653/v1/P16-1162}{https://doi.org/10.18653/v1/P16-1162}.

\bibitem[{Shen et~al.(2008)Shen, Xu, and Weischedel}]{shen2008new}
Libin Shen, Jinxi Xu, and Ralph~M Weischedel. 2008.
\newblock A new string-to-dependency machine translation algorithm with a
  target dependency language model.
\newblock In {\em ACL\/}. pages 577--585.

\bibitem[{Shi et~al.(2016)Shi, Padhi, and Knight}]{D16-1159}
Xing Shi, Inkit Padhi, and Kevin Knight. 2016.
\newblock \href{http://aclweb.org/anthology/D16-1159}{Does string-based neural
  mt learn source syntax?}
\newblock In {\em Proceedings of the 2016 Conference on Empirical Methods in
  Natural Language Processing\/}. Association for Computational Linguistics,
  pages 1526--1534.
\newblock
  \href{http://aclweb.org/anthology/D16-1159}{http://aclweb.org/anthology/D16-1159}.

\bibitem[{Stahlberg et~al.(2016)Stahlberg, Hasler, Waite, and Byrne}]{P16-2049}
Felix Stahlberg, Eva Hasler, Aurelien Waite, and Bill Byrne. 2016.
\newblock \href{https://doi.org/10.18653/v1/P16-2049}{Syntactically guided
  neural machine translation}.
\newblock In {\em Proceedings of the 54th Annual Meeting of the Association for
  Computational Linguistics (Volume 2: Short Papers)\/}. Association for
  Computational Linguistics, pages 299--305.
\newblock
  \href{https://doi.org/10.18653/v1/P16-2049}{https://doi.org/10.18653/v1/P16-2049}.

\bibitem[{Sutskever et~al.(2014)Sutskever, Vinyals, and
  Le}]{sutskever2014sequence}
Ilya Sutskever, Oriol Vinyals, and Quoc~V Le. 2014.
\newblock Sequence to sequence learning with neural networks.
\newblock In {\em Advances in neural information processing systems\/}. pages
  3104--3112.

\bibitem[{Tu et~al.(2017{\natexlab{a}})Tu, Liu, Lu, Liu, and Li}]{Tu:2017:TACL}
Zhaopeng Tu, Yang Liu, Zhengdong Lu, Xiaohua Liu, and Hang Li.
  2017{\natexlab{a}}.
\newblock Context gates for neural machine translation.
\newblock {\em Transactions of the Association for Computational Linguistics\/}
  5:87--99.

\bibitem[{Tu et~al.(2017{\natexlab{b}})Tu, Liu, Shang, Liu, and
  Li}]{tu_etal:17}
Zhaopeng Tu, Yang Liu, Lifeng Shang, Xiaohua Liu, and Hang Li.
  2017{\natexlab{b}}.
\newblock Neural machine translation with reconstruction.
\newblock In {\em Proceedings of AAAI 2017\/}. pages 3097--3103.

\bibitem[{Tu et~al.(2016)Tu, Lu, Liu, Liu, and Li}]{P16-1008}
Zhaopeng Tu, Zhengdong Lu, Yang Liu, Xiaohua Liu, and Hang Li. 2016.
\newblock \href{https://doi.org/10.18653/v1/P16-1008}{Modeling coverage for
  neural machine translation}.
\newblock In {\em Proceedings of the 54th Annual Meeting of the Association for
  Computational Linguistics (Volume 1: Long Papers)\/}. Association for
  Computational Linguistics, pages 76--85.
\newblock
  \href{https://doi.org/10.18653/v1/P16-1008}{https://doi.org/10.18653/v1/P16-1008}.

\bibitem[{Wang and Chang(2016)}]{wang2016graph}
Wenhui Wang and Baobao Chang. 2016.
\newblock Graph-based dependency parsing with bidirectional lstm.
\newblock In {\em Proceedings of ACL\/}. volume~1, pages 2306--2315.

\bibitem[{Wu et~al.(2016)Wu, Schuster, Chen, Le, Norouzi, Macherey, Krikun,
  Cao, Gao, Macherey et~al.}]{wu2016google}
Yonghui Wu, Mike Schuster, Zhifeng Chen, Quoc~V Le, Mohammad Norouzi, Wolfgang
  Macherey, Maxim Krikun, Yuan Cao, Qin Gao, Klaus Macherey, et~al. 2016.
\newblock Google's neural machine translation system: Bridging the gap between
  human and machine translation.
\newblock {\em arXiv preprint arXiv:1609.08144\/} .

\bibitem[{Zhou et~al.(2015)Zhou, Zhang, Huang, and
  Chen}]{zhou-zhang-huang-chen:2015:ACL}
Hao Zhou, Yue Zhang, Shujian Huang, and Jiajun Chen. 2015.
\newblock \href{http://www.aclweb.org/anthology/P15-1117}{A neural
  probabilistic structured-prediction model for transition-based dependency
  parsing}.
\newblock In {\em Proceedings of the 53rd Annual Meeting of the Association for
  Computational Linguistics and the 7th International Joint Conference on
  Natural Language Processing (Volume 1: Long Papers)\/}. Association for
  Computational Linguistics, Beijing, China, pages 1213--1222.
\newblock
  \href{http://www.aclweb.org/anthology/P15-1117}{http://www.aclweb.org/anthology/P15-1117}.

\end{thebibliography}
\bibliographystyle{acl_natbib}

\end{document}